\documentclass[letterpaper,twocolumn,10pt]{article}
\usepackage{usenix-2020-09}

\usepackage{tikz}
\usepackage{amsmath}

\usepackage{filecontents}
\usepackage{booktabs}
\usepackage{comment}
\usepackage{inconsolata}

\usepackage[labelfont=bf]{caption}
\usepackage{subcaption}
\usepackage{hyperref}

\newcommand{\ignore}[1]{}
\newcommand\todo[1]{\textcolor{red}{[[#1]]}}

\makeatletter
\renewcommand{\sectionautorefname}{Section}
\renewcommand{\subsectionautorefname}{Section}
\renewcommand{\subsubsectionautorefname}{Section}
\makeatother

\newcommand{\name}{PyGraph}              
\newcommand{\tool}{\name{}}                 
\newcommand{\cuda}{CUDA}
\newcommand{\cg}{CUDA Graph}
\newcommand{\cgacr}{CG}

\newcommand{\pt}{PyTorch2}
\newcommand{\ptcg}{PyTorch2-CG}
\newcommand{\ptncg}{PyTorch2-No-CG}
\newcommand{\td}{Torch Dynamo} 
\newcommand{\ti}{Torch Inductor}
\newcommand{\fg}{FxGraph}

\newcommand{\optA}{\cg{}-aware Code Transformation}
\newcommand{\optAacr}{CGCT}
\newcommand{\optB}{Parameter Indirection}
\newcommand{\optBacr}{PI}
\newcommand{\optC}{Selective \cg{}s}
\newcommand{\optCacr}{SCG}
\newcommand{\dyn}{Dynamo}
\newcommand{\ind}{Inductor}

\usepackage[frozencache,cachedir=.]{minted}

\setminted{linenos=true,autogobble,xleftmargin=15pt}

\newenvironment{pycode}
{\minted[fontfamily=cmtt,escapeinside=@@,fontsize=\scriptsize]{python}}
{\endminted}

\usepackage{tikz}
\newcommand{\circled}[1]{\tikz[baseline=(char.base)]{
            \node[shape=circle,draw,inner sep=0.5pt] (char) {#1};}}
\newcommand{\mycircled}[1]{\circled{\textbf{#1}}}

\newcommand{\runheading}[1]{\noindent\underline{\textbf{#1:}}}

\newcommand{\ie}{i.e.,}
\newcommand{\eg}{e.g.,}

\newenvironment{mybullet}{\begin{list}{$\bullet$}
{\setlength{\topsep}{0mm}\setlength{\itemsep}{0mm}
\setlength{\parsep}{0mm}
\setlength{\listparindent}{\parindent} 
\setlength{\itemindent}{0mm}\setlength{\partopsep}{0mm}
\setlength{\labelwidth}{-2mm}
\setlength{\leftmargin}{0mm}}}{\end{list}}

\usepackage{multirow}

\makeatletter
\renewcommand{\sectionautorefname}{\S\@gobble}
\renewcommand{\subsectionautorefname}{\S\@gobble}
\renewcommand{\subsubsectionautorefname}{\S\@gobble}
\makeatother

\begin{document}

\date{}

\title{\Large \bf \tool{}: Robust Compiler Support for CUDA Graphs in PyTorch} 

\author{
{\rm Abhishek Ghosh}\\
Indian Institute of Science
\and
{\rm Ajay Nayak}\\
Indian Institute of Science
\and
{\rm Ashish Panwar}\\
Microsoft Research India
\and
{\rm Arkaprava Basu}\\
Indian Institute of Science
} 

\maketitle

\begin{abstract}

Machine learning (ML) workloads launch hundreds to thousands of short-running GPU kernels per iteration. With GPU compute throughput growing rapidly, CPU-side launch latency of kernels is emerging as a bottleneck. 
CUDA Graphs promise to address this by replaying a set of kernels with a single dispatch of the graph, removing per-kernel launch costs. However, CUDA Graphs remain surprisingly difficult to deploy correctly and efficiently.

We present \tool{} --- a compiler framework to maximize the coverage and benefits of \cg{}s for ML workloads. 
It introduces three novel optimizations: it applies automatic code transformations to make ML applications amenable to \cg{}s; it eliminates the parameter copy overheads for kernels executing in \cg{}s, and it selectively deploys \cg{}s guided by a cost–benefit analysis. 
For 25 ML workloads from TorchBench, HuggingFace, and TIMM, \tool{} more than doubles the benefit from deploying \cg{} compared to the most popular and widely used ML compiler, PyTorch2.
\tool{} is built atop \pt{}’s compilation framework and requires no programmer intervention.


\end{abstract}
\section{Introduction}
\label{sec:intro}


The demand for GPU compute continues to skyrocket due to a surge in ML applications~\cite{choi:osdi.22,sarathi-serve,sarathi-short, mononn,usher, parrot,InfiniGen,mltraining-gpu-shortage, orion}. 
Catering to this growing demand through hardware enhancement \textit{alone} is challenging in the post-Moore era of slowing transistor technology~\cite{post-moore-data-center}. 
Consequently, hardware vendors increasingly rely on hardware-software co-design to improve efficiency. 
For example, NVIDIA continues to introduce new performance features in GPUs that need software enablement. 
Recent examples include scoped synchronization ~\cite{scord, iguard, scopeadvice} that allow programmers to selectively enable faster synchronization, tensor cores for fast matrix multiplication~\cite{tensor-core}, lower- and mixed-precision computation~\cite{mixedPrecisionTraining}, and \cg{}s~\cite{cudagraph, zheng:grape:23} for improving GPU utilization.
Surprisingly, GPUs are often underutilized despite high demand~\cite{pod}.
Several hyperscalers, including Alibaba and Azure, report GPU utilization typically below $50\%$~\cite{alibaba-low-gpu-util, MLaaS,azure-low-gpu-util, philly}.
A major contributing factor is that CPUs cannot submit kernels fast enough to keep modern GPUs busy.
An ML application typically invokes hundreds to thousands of GPU kernels per iteration, and each kernel launched from the CPU adds $5$-$10$ microseconds of latency \cite{launch-latency}.
As a result, GPUs stall when they finish executing kernels faster than the CPU can dispatch subsequent ones.
This problem is amplified by recent trends in distributed ML, particularly Tensor Parallelism (TP)~\cite{megatron-lm}.
TP shards model layers across GPUs, reducing the runtime of per-GPU compute kernels while introducing additional kernel launches for inter-GPU collective operations. Together, these effects push launch latency deeper into the critical path. 

\cg{}s can mitigate the kernel launch bottleneck.
It allows software to \textit{capture} kernel launches in a directed acyclic graph -- a \cg{}.
The CPU then launches the graph instead of individual kernels. 
This can significantly speed up applications that repeatedly execute the same set of kernels (\eg{} training and inference). 
While useful in theory, \cg{}s come with several constraints.
For example, the parameters to kernel are hardcoded during a graph's capture. 
The same parameter values are passed to kernels during replays posing challenge to applications that compute over different data (tensors) across invocations.
A \cg{} is also prohibited from using GPU-synchronizing operations.


Effectively harnessing capabilities of \cg{} is further made challenging by the fact that most ML applications are written in high-level frameworks such as PyTorch~\cite{pytorch} and TensorFlow~\cite{tensorflow}, and are semantically distant from their execution on the hardware.
A key challenge, thus, is to effectively bridge the gap between the high-level program semantics and the low-level idiosyncrasies of hardware features. 


Our goal is to enable high-level programs, \eg{} in PyTorch, to effectively harness \cg{}s in modern GPUs without needing programmer intervention. 
Upon qualitatively and quantitatively analyzing many tens of PyTorch programs, we observed three key shortcomings that either prevent the deployment of \cg{}s or fail to harness its benefits.

\noindent\textbf{\circled{1} \cg{}-oblivious programs:}
Many important ML programs, \eg{} in TorchBench~\cite{torchbench}, are \emph{not} written with \cg{} in mind.
For example, it is common for programs, such as the text-to-image generator model DALLE-2~\cite{dalle}, to allocate a tensor on the CPU's DRAM (say, address \texttt{x}) and then copy its content to GPU memory during computation, \eg{} in the \texttt{forward} pass.
A \cg{} captured for such computation would hardcode the address of the DRAM-resident tensor (\texttt{x}).
Its content must then be copied to GPU memory during the graph's replay.
However, the host program may de-allocate the CPU tensor (\texttt{x}) after the graph's capture, causing the application to crash intermittently during replays.
Thus, \cg{} must be avoided in such programs. 
We notice similar pattern in other programs \eg{} Vision Mask R-CNN~\cite{maskrcnn}, Torch Multi-modal Clip~\cite{torchmultimodal, clip}.

Further, in many cases  
one or more kernel parameters are passed as \textit{scalar} variables \eg{} attention kernel in speech transformer~\cite{speech}.
A \cg{} captured with such kernels would record the value of the scalar as the kernel parameter. 
If the variable's value should change across kernel invocations, it will receive \textit{stale} values during graph replays.
Thus, \cg{} must be avoided in such cases.


While a single or a few lines of code changes can unlock hundreds of kernels being captured in \cg{}{}s, the challenge is to automatically find such opportunities and update the code accordingly.  
It is unrealistic to expect developers to continue evolving their programs to make them friendlier to new hardware features.
Thus, we posit that frameworks, such as \pt{}, JAX~\cite{jax}, must step up to effectively harness advanced hardware features without burdening developers. 

\noindent\textbf{\circled{2} Parameter copy overheads in replaying \cg{}s:}
By design, \cg{}s captures GPU kernel parameters, including addresses of tensors \emph{by value}. 
However, different invocations of the same kernel across a graph's repeated replays often need to operate on different data. 
A simple strategy to enable this is to replace mutable parameters with \textit{placeholder} parameters. 
Before each replay, actual parameter values for a given invocation are copied onto its placeholders. 


While this strategy allows correct replay of graphs and is employed by widely used \pt{} ML compiler, it adds the overhead of copying the parameter (data) into the placeholder in the critical path of every graph replay.
We empirically found overheads of copying parameters constitute up to 24\% of a \cg{}'s execution time, particularly when operating on large tensors (\autoref{sec:bg}).
Thus, there is a need to correctly deploy \cg{} but without slowing down graph replays. 

\noindent\textbf{\circled{3} \cg{}s may hurt performance:} While \cg{} is a performance feature, we discovered that the overhead of replaying them can sometimes outweigh the benefits. 
Besides the latency of copying kernel parameters (tensors), the overhead of reusing/garbage collecting \cg{}'s memory after replay and reset of the GPU's random-number generator state before each replay adds latency to the critical path.
In our analysis, nearly a quarter of the total \cg{}s degrade performance (up to 397\%). 
While a performance feature is not expected to speed up every application, it must not severely slow down many applications.

\noindent\underline{\textbf{\tool{}:}} Driven by these observations, we created a \cg{}-aware compilation framework for PyTorch programs. 
\tool{} embodies three key optimizations.

\textbf{\circled{1}} It employs \cg{}-aware code transformations such as hoisting memcopies, changing data placement for a wider deployment of \cg{}s. 
It analyzes the intermediate representation (IR) of programs to root cause the failures to capture kernel(s) in a graph (\eg{} scalar variables).
It then updates the IR to emulate changes in the source that would have enabled \cg{}. 
\textbf{\circled{2}} It introduces an additional level of indirection in kernel parameters to convert tensor copies for parameters to \textit{pointer} copies.
While a tensor can be thousands of bytes, a pointer is only 8 bytes.
For JIT-compiled kernels, \tool{} uses Triton compiler~\cite{triton} to automatically update kernels to de-reference their indirect-ed parameters.
For vendor-supplied immutable kernels (\eg{} from cuBLAS), it introduces prelude kernels in the graph that perform de-reference of indirect-ed parameters on behalf of immutable kernels.
\textbf{\circled{3}} Finally, it adds automated profiling during graph capture (slow path) to analyze the cost-benefit of deploying \cg{}s. 
This guides \tool{} to \textit{selectively} deploy \cg{} during replay (fast path) only when beneficial. 
The judicious deployment of \cg{}s not only enables better end-to-end performance but, crucially, ensures that no application slows down. 
\tool{} is integrated into \pt{} and requires \textit{no} program modifications. 

On a diverse set of $25$ ML applications, \tool{} provides geomean speed up of 29\% over \pt{}'s \cg{} feature (and up to 3.36$\times$) on an H100 GPU.   
Further, we studied \tool{}'s efficacy under distributed ML workloads by spreading a subset of applications over four GPUs using tensor parallelism. 
There, \tool{} speeds up applications up to $3.56\times$ over \pt{}'s \cg{} and by 75\% on average. 

The key contributions of our work are as follows:
\begin{mybullet}
    \item We demonstrate subtle programming idiosyncrasies in popular PyTorch programs, coupled with the constraints of \cg{}s, limit wider deployment of \cg{}s.
    \item We show that overheads due to copying kernel parameters can be limited by introducing an additional layer of indirection in passing the parameters to kernels. 
    \item We demonstrate the need to selectively deploy \cg{}s to achieve better end-to-end application performance. 
    \item We create \tool{} atop \pt{}'s compilation framework that achieves the above-mentioned optimizations without needing any manual modifications. 
\end{mybullet}

\section{Background and Motivation}
\label{sec:bg}
\label{sec:motivation}

This section elaborates on how \cg{}s have become indispensable for modern GPU platforms, along with the challenges involved in maximizing their utilization. We use the terms \cg{}(s) and \cgacr{}(s) interchangeably.

\subsection{The need for \cg{}s}
GPUs have become the workhorse accelerators for ML applications since they deliver massive compute throughput. 
NVIDIA’s GPU compute scaling highlights this: peak FP16 tensor throughput jumped from 312~TFLOP/s on A100~\cite{A100} to \(\approx\)1~PFLOP/s on H100~\cite{H100}, and \(\approx\)4~PFLOP/s on B200~\cite{B200}. 
In contrast, CPU frequency has stagnated, and the number of cores has only doubled during that period. 
At the same time, ML workloads have expanded dramatically, with each iteration often launching hundreds to thousands of kernels.
Further, modern deployments increasingly shard a model across multiple GPUs.
As GPU throughput improves and models scale out, kernels become smaller and faster. 

A typical CUDA kernel launch takes $\sim$5-10 microseconds~\cite{launch-latency}, while many kernels now run for only a few microseconds. This turns CPU launch overhead into a significant fraction of the total time. 
Reducing or amortizing this overhead is therefore critical to achieving high GPU utilization. 
For example, in DALLE2~\cite{dalle} (inference, batch size 1), a segment of computation issues over 740 kernels whose combined GPU execution time is only 3.4 milliseconds on an H100 GPU, while the end-to-end time is 14 milliseconds—implying that 75\% of the total latency is due to CPU launch delays.

\begin{figure}[t!]
\centering
\includegraphics[width=.85\linewidth]{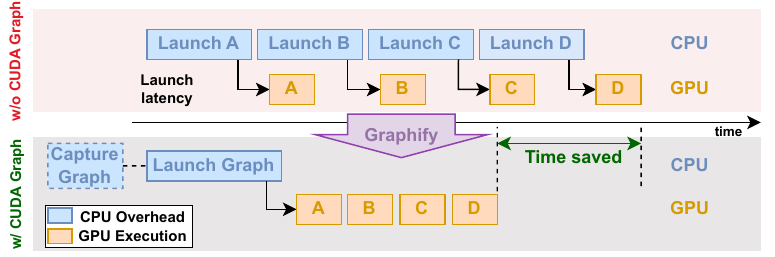}
\caption{\cg{} reducing CPU launch overheads.}
\label{cudagraph-launch-latency-squash}
\end{figure}

A \cgacr{} can address this bottleneck by obviating the need to individually launch each kernel from the CPU. 
The top part of \autoref{cudagraph-launch-latency-squash} shows traditional kernel launches, while the lower part shows steps involved in \cgacr{}. 
Kernels are captured as part of a DAG (\cgacr{}) once. 
The graph is launched from the CPU, and internally, the GPU hardware launches \cgacr{}'s constituent kernels without CPU intervention. 
Applications that repeatedly launch the same set of kernels, \eg{} inference tasks, can benefit substantially by capturing the graph once and replaying it many times.  

\subsection{Deploying \cg{} and its challenges}

\cuda{} supports automatic capture of \cgacr{}s through \emph{stream capture} mechanism. 
The developer designates a capture region.
The runtime automatically records all GPU operations (\eg{} kernel launch, memcopy) issued within the region, producing a computation DAG. 
Once capture completes, the developer instantiates the graph and launches it, causing the GPU to \emph{replay} the recorded GPU operations. 
Owing to its simplicity, stream capture is the dominant way developers use \cgacr{}s. 
However, even this approach requires care: the runtime records kernel parameters \emph{by value}. 
If the application replays the \cgacr{} without updating parameters that change across iterations, execution proceeds with stale arguments—leading to silent data corruption, incorrect results, or crashes. 
Developers must therefore identify which values remain stable across replays and update any mutable parameters in the graph before launching it. 
Furthermore, \cuda{} does not permit capturing synchronous copies (\texttt{cudaMemcpy}), memory allocation (\texttt{cudaMalloc}) or any other API that implicitly synchronizes the device~\cite{cudagraph-doc}. 
These constraints force developers to rethink data movement and memory management, making it difficult to incorporate \cgacr{}s—especially in legacy codebases. 

We posit that compilers and \textit{not} the programmers should reason about program structure, automatically satisfy graph constraints and judiciously apply optimizations.
This is the only way to harness advanced hardware features without burdening the developers. 
Unfortunately, existing compilers fall well short. 
To substantiate, we investigate PyTorch2~\cite{pytorch2}---the most popular and widely used ML compiler. 

\begin{figure}[tb!]
    \centering
    {\fontfamily{zi4}\selectfont
    \begin{pycode}
    import torch 
    import numpy as np 
    
    class ScaledDotProductAttention(nn.Module): 
      def __init__(self, temperature, attn_dropout): 
        ...  
        self.temperature = temperature 
        ... 
    
      def forward(self, q, k, v, mask=None): 
        attn = torch.bmm(q, k.transpose(1, 2)) 
        # Divide operation below with a numpy scaler on CPU
        attn = attn / @\textcolor{red}{self.temperature}@ # @\textcolor{brown}{offending line}@
        ...
    
    class MultiHeadAttention(nn.Module): 
      def __init__(self, n_head, d_model, d_k, d_v, dropout): 
        ...
        @\textcolor{red}{temp = np.power(d\_k, 0.5)}@ # @\textcolor{brown}{src of offending operand}@
        self.attn = ScaledDotProductAttention(
            temperature=@\textcolor{red}{temp}@, attn_dropout=dropout) 
        ...   
    
      def forward(self, q, k, v, mask=None): 
        ... 
        output, attn = self.attn(q, k, v, mask=mask)    
        ... 
    
    module = # Initialize the speech_tranformer model 
    module = torch.compile(module) 
    result = module(inputs)
    \end{pycode}
    }
    \caption{Simplified code snippet from speech transformer model with CPU-resident scalar (marked red).}
    \label{fig:cpu-tensor-example}
\end{figure}

\subsection{\cg{}s are poorly harnessed}
We analyze over 60 common ML and HPC workloads to identify key impediments to effective \cgacr{} deployment. 
First, many programs fail to use \cgacr{}s because their implementations are unaligned to its requirements.
Second, \cgacr{}'s deployment strategy today adds significant but avoidable overheads. 
Further, even when \cgacr{}s can be deployed, their deployment is not always beneficial — yet current frameworks deploy them blindly. We present a detailed analysis below.

\subsubsection{Obstacles in harnessing \cg{}s}

We observe that \cg{}s  are often not compatible with the way many ML workloads are written. For example, when a kernel is captured into a \cgacr{}, all its arguments are recorded by value and baked into the graph. If those values change across iterations, replaying the captured graph would produce incorrect results. Consequently, a compiler would forgo \cgacr{} deployment in these cases for correctness.

We highlight one such example with a simplified PyTorch code snippet of the speech transformer model~\cite{speech} (\autoref{fig:cpu-tensor-example}). In the constructor of \texttt{MultiheadAttention} (line 19), the variable \texttt{temperature} is assigned the value returned by a \texttt{numpy} method~\cite{numpy}. Note that \texttt{temperature} hosts a \textit{scalar} value and is allocated on the CPU DRAM. This variable is passed to the constructor of \texttt{ScaledDotProductAttention} (line 21) and ultimately used to update \texttt{attn} in its forward method (line 13). This update of \texttt{attn} is performed by a GPU kernel whose parameters are \texttt{attn} and \texttt{temperature}. While the former is the address to a GPU tensor, the latter is a \textit{scalar} variable. If this kernel is captured as part of a \cg{}, the scalar value of the parameter \texttt{temperature} will be hardcoded as a kernel parameter. Later, if the captured graph is replayed, the kernel would execute with a stale value of \texttt{temperature}.

We discovered other subtleties in PyTorch programs that limit \cgacr{} deployment. A common case is when applications allocate a tensor on the DRAM (at address \texttt{x}) and copy its contents to GPU memory at the start of each iteration. A captured \cgacr{} would hardcode the DRAM address \texttt{x}.
However, CPU may de-allocate that tensor after capture, leaving the address invalid when the graph is replayed.

\begin{figure}[t!]
\centering
\includegraphics[width=.65\linewidth]{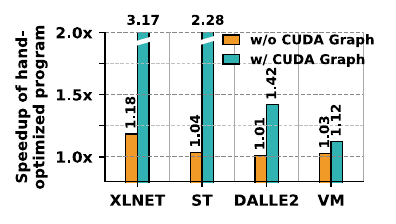}
\caption{Performance impact of moving tensors from CPU (unoptimized programs) to GPU (optimized programs).}
\label{fig:gpu-tensor-speedup}
\end{figure}

\noindent
\textbf{Quantifying the opportunity cost:} 
The issues above stem from the fact that most ML programs are not written to meet \cgacr{} constraints. 
In pre–\cgacr{} era, choices about where to allocate tensors—CPU vs.\ GPU—had minimal impact on performance, so developers had little reason to care. With the introduction of \cgacr{}s, these decisions suddenly matter: seemingly innocuous placement choices can make a dramatic impact on performance because even a single CPU-resident tensor can prevent the use of \cgacr{}s for many kernels. To demonstrate this, we manually rewrote small portions of several PyTorch models—XLNetLMHeadModel (XLNET)~\cite{xlnet}, SpeechTransformer (ST)~\cite{speech}, DALLE2~\cite{dalle}, and Vision Mask R-CNN (VM)~\cite{maskrcnn}—specifically targeting code regions whose structure prevented \cg{} deployment. The edits were minor (e.g., relocating a CPU-allocated tensors to the GPU) but made the code \cgacr{}-compatible. \autoref{fig:gpu-tensor-speedup} reports the results. (All results use inference batch size 1). In the absence of \cgacr{}s, optimizing tensor placement had a modest impact on performance, averaging 6.29\% and reaching up to 18\% for XLNET. However, the same code changes lead to a dramatic speedup when using \cgacr{}s. For example, XLNET and ST sped up by $3.17\times$ and $2.28\times$! 

Furthermore, scaling out ML models to multiple GPUs makes \cgacr{}s more critical for performance. For instance, in the XLNET model (inference at batch size 128), tensor-parallelism over four GPUs brings 48 additional collective kernels and reduces the runtime of more than 250 kernels by 50–75\%. 
However, a few variable allocations in CPU DRAM render more than 450 kernels ineligible for capture in \cgacr{}s, leaving significant performance on the table. 

\begin{figure}[t!]
\centering
\begin{subfigure}{0.49\linewidth}
    \centering
\includegraphics[width=\linewidth]{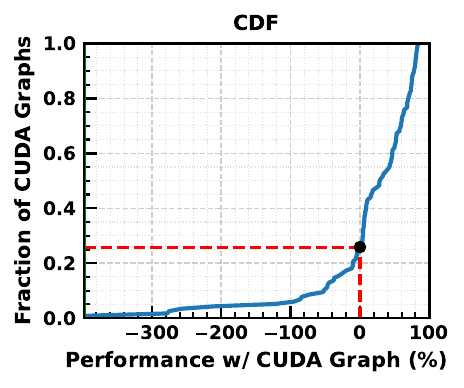}
    \caption{\scriptsize  $25\%$ of the total \cg{}s hurt performance in TorchBench.}
    \label{subfig:cdf-frac-cg-against-cg-perf}
\end{subfigure}
\hfill
\begin{subfigure}{0.49\linewidth}
    \centering
    \includegraphics[width=\linewidth]{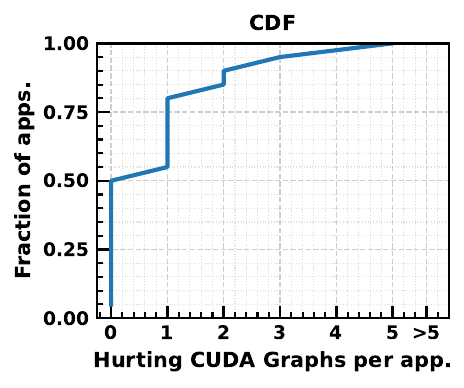}
    \caption{\scriptsize About 50\% of the applications contain \cg{}(s) that hurt performance.}
    \label{subfig:cdf-frac-apps-against-hurting-cg}
\end{subfigure}
\caption{Cost-benefit analysis of \cg{}s.}
\label{selective-disable-motivation-stats}
\end{figure}

\subsubsection{Cost-benefit analysis of \cg{}s}
We also noticed that even when \cgacr{}s can be deployed, they do not always improve performance. In fact, counterintuitively, we observed that \cgacr{}s also hurt performance in many cases. This is unexpected; \cgacr{}s are designed solely as a performance feature and should not slow down even a single application. To understand this behavior, we quantitatively analyze 116 \cgacr{}s drawn from 63 PyTorch applications in TorchBench~\cite{torchbench}.

We make two key observations. First, \autoref{subfig:cdf-frac-cg-against-cg-perf} shows the CDF of performance changes when enabling \cgacr{}s across all applications, capturing both cases where \cgacr{}s help (right side of 0 on the x-axis) and where they hurt (left side of 0). Overall, 25\% of the graphs (29 of 116) degrade performance. A deeper inspection reveals that the dominant source of this slowdown is the latency of copying kernel parameters (data) into the static placeholders (addresses) during a graph's replay.

The reason this copying is required is fundamental to how \cgacr{}s operate: every kernel parameter is captured by value in a \cg{}, whereas different invocations of a kernel typically operate on different data. 
A simple strategy to manage this scenario is to replace mutable parameters with \textit{place-holder} arguments. 
Before each replay, the runtime must overwrite these placeholders with the current input values of all mutable arguments.
This copy happens on the critical path of every replay and scales with the number of kernels and the size of their argument lists, making it a nontrivial overhead. 
For instance, in \cg{}s of DALLE2 and NVIDIA Deep Recommender, parameter copying accounts for 24\% and 17\% of their end-to-end execution time, respectively. 
These overheads substantially erode the benefits of \cgacr{} deployment and can even lead to net slowdowns.

Beyond parameter copy, replay of a \cgacr{} suffers from other fundamental overheads. 
\cgacr{}s cannot encode lifetimes of CPU-side objects.
A language runtime that performs garbage collection, e.g., Python, must separate \cgacr{} outputs from the host objects that originally produced them.
The runtime snapshots only the GPU-side metadata of \cgacr{} outputs (such as data pointers and layout) and discards the CPU-side tensor objects.
This is necessary for the GPU memory to be safely reused across \cgacr{} replays without memory bloat.
However, on each replay, fresh CPU objects must be reconstructed from the saved metadata, adding further latency.


Our second insight is that many applications consist of both useful and harmful \cgacr{}s. For example,~\autoref{subfig:cdf-frac-apps-against-hurting-cg} shows the CDF of 20 applications that contain one or more \cgacr{}s.
Of these, nearly half of them exhibit a mix of performance-aiding and performance-hurting \cgacr{}s. For a specific example, in DALLE2 (inference, batch size 1), one of the captured \cgacr{} executes in 75 microseconds, while the kernel in it when run without \cgacr{} takes 70 microseconds—implying an overhead of 7\% in \cgacr{} deployment. At the same time, a different \cgacr{} in the same workload reduces the runtime of 743 kernels to 3.4 milliseconds, compared to 14 milliseconds without \cgacr{}s, resulting in more than 75\% performance gain.
Unfortunately, \cgacr{} today are blindly deployed  without a cost-benefit analysis. 

In summary, these expose a deeper problem: today’s ML software stack lacks both the semantic understanding needed to effectively harness \cg{}s and a cost model to decide when they are worthwhile. 
Applications often miss crucial opportunities to use \cgacr{}s, or deploy them where they hurt performance. 
It begs for a compiler-driven approach to automatically enforce graph compatibility, reason about costs, and maximize the effective use of \cgacr{}s.


\section{\tool{}: Goals and Optimizations}
\label{sec:design}

Motivated by the analysis, we present the goals of \tool{}, a framework that enables effective use of \cg{}s in ML applications. We then outline the optimizations that help \tool{} achieve the goals.


\noindent\textbf{Goals:} 
\circled{\textbf{1}} Expand the set of GPU kernels that can be captured in \cgacr{}s.
\circled{\textbf{2}} Minimize runtime overheads introduced during \cgacr{} replay, especially those stemming from parameter copy.
\circled{\textbf{3}} Selectively deploy \cgacr{}s only when they improve end-to-end performance.
\circled{\textbf{4}} Achieve all of the above without requiring modifications or developer's intervention.

\runheading{\optA{} (\optAacr{})}
A large fraction of ML applications fail to utilize \cg{}s due to the presence of constructs—synchronous memory copies, CPU-resident scalars or tensors—that inhibit \cgacr{} capture. These issues are often obscured by high-level frameworks and can easily escape a programmer's attention.

\tool{} addresses this by restructuring the program such that large sequences of kernels become eligible for graph capture, without altering application semantics. 
It first analyzes the compiler’s intermediate representation (IR) to identify code regions where \cgacr{} capture fails and the root cause of the failure. 
It then performs targeted IR rewrites that make these regions compatible with \cgacr{} capture. 
Conceptually, \tool{} ensures that variables and operations in a candidate region satisfy the constraints imposed by \cgacr{}s: computations and necessary inputs are placed on the GPU, and incompatible operations are hoisted prior to candidate graph regions. For example, if a scalar parameter prevents graph capture, \tool{} alters the type of parameter as a GPU-resident tensor.  During replays, the appropriate value of a parameter is copied into the tensor’s placeholder, enabling safe capture of the surrounding kernels. Likewise, if a DRAM-to-GPU memory copy prevents \cgacr{} deployment, \tool{} hoists the copy above the sequence of kernels targeted for \cgacr{} capture. This ensures that the copy is not captured as part of the \cgacr{} and, thus, need not be replayed. 



\runheading{\optB{} (\optBacr{})}
A dominant source of overhead in a \cgacr{} replay is the copying of kernel parameters (data) into the placeholders that were recorded during graph capture.
We eliminate this cost by removing data copies from the replay path entirely: instead of copying potentially large parameter data, we introduce a layer of indirection.
Consequently, the replay updates only the corresponding pointers (eight bytes per parameter) rather than the data itself, which can be thousands of bytes. 
This optimization rests on a simple observation: with the input data changing on each replay, as long as a kernel can read from these new data tensors—different from those it was captured with—the \cgacr{} execution remains valid, without needing a parameter copy.


\begin{figure}[t!]
\begin{subfigure}{\linewidth}
\centering
\includegraphics[width=\linewidth]{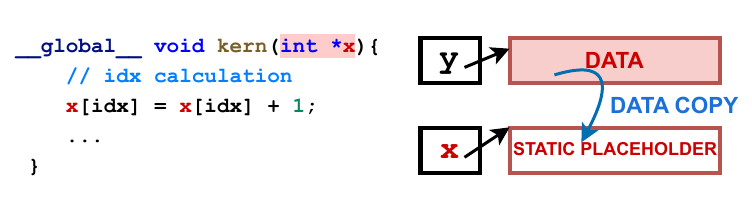}
\caption{Baseline behavior in current ML frameworks.}
\label{subfig:before}
\end{subfigure}
\begin{subfigure}{\linewidth}
\centering
\includegraphics[width=\linewidth]{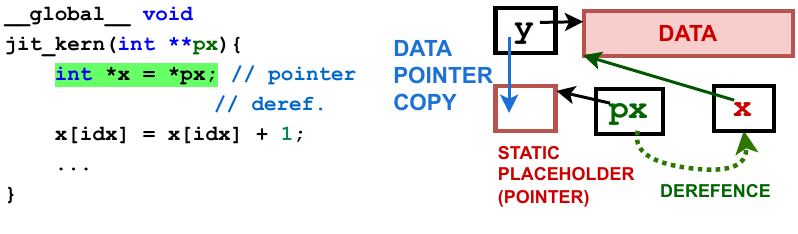}
\caption{After indirection.}
\label{subfig:after}
\end{subfigure}
\caption{Converting data (parameter value) copy to pointer copy through indirection in \optB{}.}
\label{derefernce}
\end{figure}

\begin{figure*}[t!]
    \centering
    \includegraphics[width=0.8\linewidth]{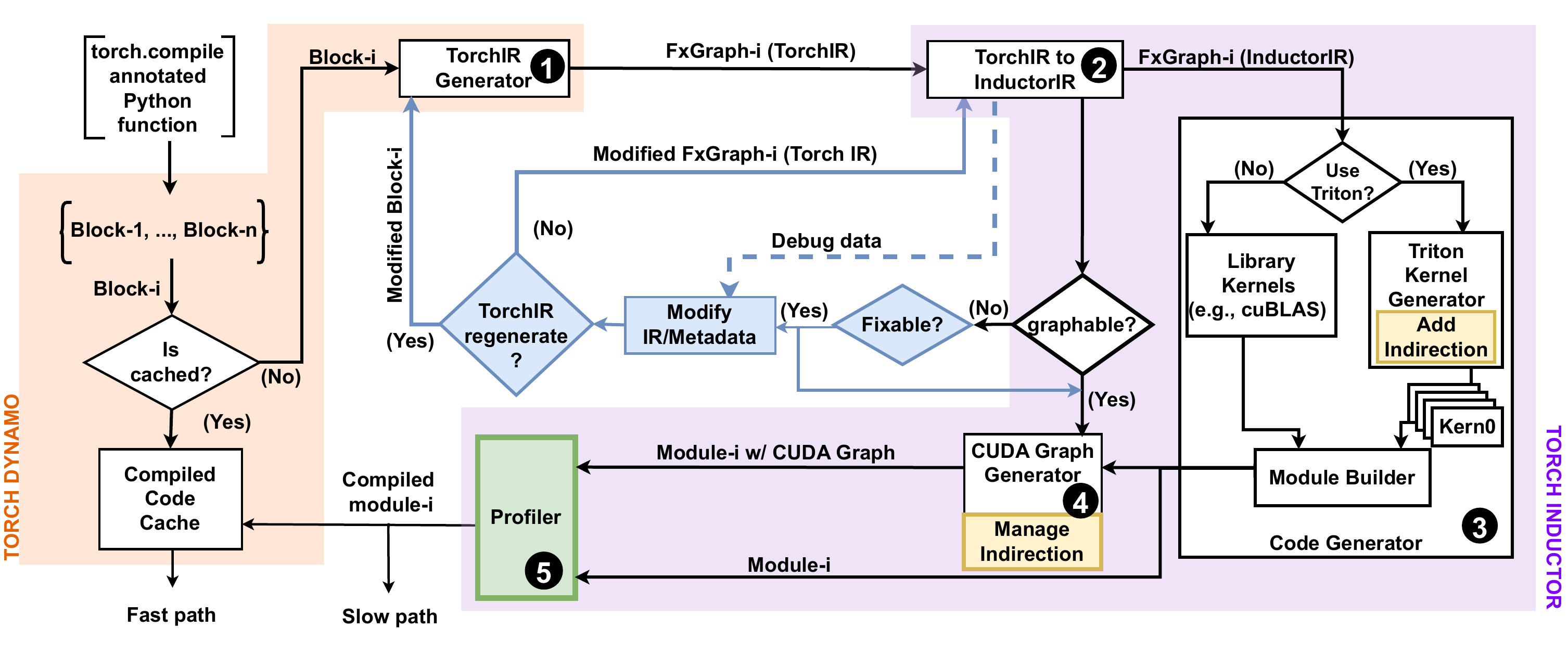}
    \caption{Implementation of \tool{} within \pt{}'s compilation framework.}
    \label{fig:impl:main}
\end{figure*}

\autoref{derefernce} illustrates this mechanism with an example.
\autoref{subfig:before} shows the parameter (data) copy performed during the replay of a \cgacr{} in the baseline. 
The kernel parameter recorded during graph capture (here, address \texttt{x}) is highlighted in red. 
During replay, the parameter for a given kernel invocation, residing at address \texttt{y} — must be copied into the memory location pointed to by \texttt{x}. \autoref{subfig:after} shows how \tool{} converts the kernel parameter to pointer-to-pointer via indirection. 
These pointer-to-pointers are then recorded during graph capture. Before each replay, the pointers to the parameter (\texttt{y}) for the given invocation of the kernel are copied onto the address (pointer-to-pointer, \texttt{px}) recorded during the graph capture.  The modified kernel then de-references these pointer-to-pointers before performing any computation (green).


Our approach relies on the ability to rewrite kernels, which is feasible for JIT-compiled kernels (e.g., Triton~\cite{triton}) but not for vendor-supplied libraries (e.g., CUTLASS~\cite{cutlass}, cuBLAS~\cite{cublas}) or pre-compiled kernels. In \autoref{sec:implementation}, we describe an alternative mechanism that extends \optBacr{} to these kernels based on recently introduced \cgacr{} management APIs~\cite{graph-management-api}, while meeting \tool{}’s goal of zero manual changes.



\runheading{\optC{} (\optCacr{})}
Since \cgacr{}s can also hurt performance (\autoref{sec:motivation}), they should not be used blindly. \tool{} employs automated profiling as part of the compilation and graph capture phase (slow path). 
Specifically, \tool{} measures the aggregate time to execute individual kernels encapsulated in a \cgacr{}. 
It also profiles the execution time of the corresponding \cgacr{} with and without \optB{}. 
The profiler then automatically chooses the best-performing configuration for the given set of kernels in an application.
This decision is made independently for each of the \cgacr{}s.  Also, note that this profiling happens during compilation and \cgacr{} capture phase (slow path) and not during the graph replay (fast path).

Taken together, these three optimizations push \cg{}s closer to their peak potential. \optAacr{} broadens capture eligibility so that substantially more of the program can benefit from graph execution. \optBacr{} removes a major replay bottleneck by eliminating parameter-copy overheads. \optCacr{} ensures that \cgacr{}s are enabled only when they deliver end-to-end gains. The result is a system that captures larger kernel regions in graphs, executes them with lower overhead, and deploys \cgacr{}s only where they improve performance.


\section{Implementing \tool}
\label{sec:implementation}
\tool{} is built into \pt{}'s compilation workflow (slow path) that is responsible for \cg{} capture, enabling its optimizations to benefit fast-path execution (replay).
While our prototype is implemented in the \pt{} compiler, its optimizations are generic and designed to address fundamental challenges of using \cgacr{}s, making them applicable to most frameworks. We begin with a brief discussion of the compilation workflow of the \pt{} compiler and then discuss how \tool{} leverages it to apply its three optimizations.

\noindent
\textbf{Compilation workflow in \tool{}:} \tool{} follows a structure commonly found in modern ML compilers—such as PyTorch2~\cite{pytorch2}, JAX/XLA~\cite{jax}.
The typical compilation pipeline consists of
\circled{1} \textit{Computation Graph Construction Layer (Frontend)}, where Python execution is symbolically traced, producing a computation graph in a high-level IR.
\circled{2} \textit{IR Lowering \& Optimization Layer (Mid-Level Pass)}, where the graph is lowered into a more uniform IR, enriched with metadata such as tensor shapes, device placement, and debug mappings.
\circled{3} \textit{Kernel Generation Layer (Backend)}, where primitive operations are either dispatched to optimized GPU libraries (cuBLAS~\cite{cublas}, cuDNN~\cite{cudnn}, CUTLASS~\cite{cutlass}) or implemented through just-in-time (JIT) kernel generators such as Triton~\cite{triton}.
\circled{4} \textit{Graph-Capture Layer}, where the final compiled module is executed, and if eligible, captured as a \cgacr{}. This layer maintains parameter placeholders and bookkeeping logic required to replay \cgacr{}s.

\tool{} adds targeted extensions to each of these layers.
\autoref{fig:impl:main} provides an overview of \tool{}’s implementation, depicting 
how it extends selected components of the \pt{}'s compilation workflow. 
The orange and purple background regions delimit the key components of the \pt{} compiler—its frontend (\td{} \cite{td}) and backend (\ti{} \cite{ti}).
Modifications for \optAacr{} are shown in blue, while those for \optBacr{} and \optCacr{} appear in yellow and green, respectively. 
Its three components—\optAacr{} (blue), \optBacr{} (yellow), and \optCacr{} (green)—extend different parts of the compilation workflow and are detailed in \autoref{sec:iirgen}, \autoref{sec:cgify}, and \autoref{sec:selector}, respectively.

\subsection{\optA{}}
\label{sec:iirgen}

\tool{} automatically finds and mitigates multiple factors that prevent deployment of \cg{}s.  For ease of exposition, we describe the \pt{} components involved in deploying \cgacr{}s first, and then \tool{}'s extensions to them.

\subsubsection{Computation Graph Construction Layer}

\td{} (\autoref{fig:impl:main}, orange block on the left) intercepts Python execution and constructs computation graphs (\fg{}s~\cite{torchfx}) consisting of TorchIR — a simplified IR that exposes Python and PyTorch operations for optimization. 
\dyn{} symbolically traces operations until it encounters one that cannot be safely traced (\eg{} input-dependent control flow).
It terminates the current \fg{}, compiles it, and splices the compiled result back into the source program via bytecode rewriting. 
It resumes tracing after the offending operation, effectively partitioning the program into blocks, each of which is lowered independently to TorchIR (\mycircled{1} in \autoref{fig:impl:main}).

\runheading{\tool{}'s extensions} 
\dyn{} is extended to regenerate TorchIR from a modified code block (\mycircled{1} in \autoref{fig:impl:main}), provided by the extensions to \ti{} described next. 

\subsubsection{IR-Lowering \& Optimization Layer}
\ti{} (\autoref{fig:impl:main}, purple region), lowers the TorchIR produced by \dyn{} into a lower-level IR comprised of primitive Aten/Prim~\cite{aten-prim} operations.
\ind{} then augments this IR with operand metadata--including tensor shapes, device placement (CPU or GPU), and debug mappings from InductorIR back to TorchIR--producing InductorIR (\mycircled{2} in \autoref{fig:impl:main}). InductorIR serves as input to subsequent phases (\autoref{sec:cgify}) that apply further optimizations and code transformations.

\ind{} also decides whether a \fg{} represented by an InductorIR instance can be captured as a \cg{}. 
If the metadata shows that an operand of a \fg{}'a node is a CPU-resident scalar, a CPU--GPU memcopy, or an input mutation, it avoids encapsulating it in a \cgacr{}.

\runheading{\tool{}'s extensions} 
To map more \fg{}s to \cgacr{}s, \tool{} enhances \ind{}’s decision routine for \cgacr{} eligibility. 
When this routine determines that an \fg{} cannot be mapped to a \cgacr{}, \tool{} inspects the cause of failure. 
If the failure is due to the presence of one or more scalar variables (\eg{} a NumPy scalar as discussed in \autoref{sec:motivation}), it uses bytecode rewrites to cast the variables as GPU tensors, effectively removing the CPU operand/operation from the graph capture path.  
\dyn{} then regenerates TorchIR for the modified block, which is re-lowered to InductorIR, enabling the \fg{} to be mapped to a \cgacr{}.

A CPU-to-GPU memcopy can also prevent \ind{} from deploying a \cgacr{}. 
Using debug metadata, \tool{} backtraces the InductorIR node to the Python object owning the offending CPU tensor and copies it to the GPU. 
This effectively changes the variable’s device placement, hoisting the memcopy into the object’s constructor so that it precedes any computation. 
As a result, memcopy gets removed from the graph-capture path, and the corresponding module can be mapped to a \cgacr{}.
Similarly, a \cgacr{} may not get deployed if the output of an InductorIR node is a CPU tensor. 
In such a case, \tool{} updates the metadata of the TorchIR node to mark the output as a GPU tensor (\ie{} setting \texttt{device=`cuda'}). 
The updated TorchIR is then re-lowered to InductorIR (\mycircled{2} in \autoref{fig:impl:main}), removing the CPU tensor from the graph-capture path and allowing the \fg{} to be mapped to a \cgacr{}.


\subsection{\optB{}}
\label{sec:cgify}

We detail how \tool{} achieves its goal of reducing parameter copy overheads through indirection. 
This needs modifying how kernels within \cg{}s obtain their parameters. 

\subsubsection{Kernel-Generation Layer}

The code generator (\mycircled{3} in \autoref{fig:impl:main}) is responsible for generating an optimized executable (module) for an \fg{}. 
It uses different frameworks/libraries depending on the target hardware. 
Here, we focus on systems with NVIDIA GPUs. 

The code generator either invokes kernels from vendor-supplied external libraries (\eg{} cuBLAS~\cite{cublas}) or generates kernels just-in-time (JIT) using frameworks such as OpenAI's Triton~\cite{triton}.  
It maintains a list of InductorIR ops (e.g., matmul, convolution) that use kernels from vendor-supplied libraries and emits direct calls to the relevant library routines.
Remaining operations are implemented just-in-time using Triton. 
For such operations, the code generator constructs a Triton function in Triton’s Python-based DSL, decorates it with \texttt{triton.jit} for JIT compilation.
It then invokes the Triton compiler to generate CUDA binaries. 
A module created for an \fg{} may contain both Triton and vendor-supplied kernels. 
After resolving all operations, the module builder assembles these components into a single, optimized module.

\runheading{\tool{}'s extensions}
Recall that the goal of \tool{} is to enable repeated replay of a captured \cgacr{} with different parameter values, but without adding the overhead of copying the parameter. 
It makes two changes to kernels to turn parameter (data) copies into pointer copies.
First, it replaces each pointer argument with a pointer-to-pointer that holds the address of the data region the kernel will access.
Second, before the kernel's computation starts, it dereferences these pointers-to-pointers, allowing the rest of the kernel to access the actual data correctly.

\tool{} employs two distinct implementation approaches to achieve the above objective, depending on whether the given kernel is generated just-in-time or is a vendor-supplied one. 
While the former can be updated during compilation, the latter is \textit{not}  -- requiring two different approaches. 

\noindent
\textbf{Enabling \optBacr{} in JIT-ed kernels:} To apply indirection to eligible Triton-generated JIT kernels without manual intervention, \tool{} identifies the parameters that must undergo indirection during the code generation phase~\footnote{Not all kernels or parameters require copying parameter data into placeholders. 
Outputs produced by earlier kernels in a \cgacr{} can be passed directly to later kernels. 
Only kernels whose parameters originate from external sources (\eg{} application inputs) require indirection. 
\tool{} selectively applies indirection only for such parameters.} and attaches this information to the \texttt{torch.jit} decorator (yellow element in \mycircled{4} in \autoref{fig:impl:main}). 
\tool{}'s custom LLVM~\cite{llvm} pass integrated into the Triton framework then retrieves this parameter list from the decorator and rewrites the PTX~\cite{ptx} code of the corresponding kernels. 
The pass updates the kernel signatures to use `pointer-to-pointer' arguments for the designated parameters. 
Further, it inserts PTX instructions at the beginning of the kernel to de-reference these pointers before using them in computation. 
The rest of the kernel code remains unchanged.

\noindent
\textbf{Enabling \optBacr{} in vendor-supplied kernels:}
The approach of rewriting kernels is not feasible for those from vendor-supplied external libraries. 
These kernels are often supplied as compiled binaries. 
To workaround this fundamental limitation, \tool{} introduces an additional \textit{prelude} kernel in the \cg{} at the start of the graph. 
The prelude kernel updates the parameters of the vendor-supplied kernel to point to actual data for a given invocation on the fly during replays. 

The prelude kernel is provided with pointer-to-pointers to the parameters of the vendor-supplied kernel, whose values change across graph replays.
It is tasked with performing the dereferencing of the pointer-to-pointer argument for the immutable vendor-supplied kernel. 
\tool{} compiles the prelude kernel on the fly using NVRTC~\cite{nvrtc}, and re-records the \cg{} with this kernel inserted before other nodes.
To achieve the effect of dereferencing, the prelude kernel must be able to update parameters to the vendor-supplied kernel on the fly during replay.  

Fortunately, CUDA runtime version 12.4, introduced new device-side APIs to update parameters of a node in \cg{} from within another kernel (\texttt{cudaGraphKernelNodeSetParam})~\cite{graph-management-api}. 
The prelude kernel uses this API to update the parameters of the vendor-supplied kernel during replay. 

The device-side APIs to update kernel parameters take three arguments: \circled{\textbf{1}} the kernel-node handle, \circled{\textbf{2}} a device-memory address (pointer) that holds the replacement pointer value, and \circled{\textbf{3}} the offset within the kernel node’s parameter buffer. 
The API de-references the input pointer and writes the resulting value into the parameter buffer at the specified offset for the given kernel node. 
This operation performs a de-reference followed by a memory write.

The next challenge is to find the offset of the argument to be updated within the kernel's parameter buffer. 
For this purpose, \tool{} uses another CUDA API \texttt{cuFuncGetParamInfo} to obtain a handle to the kernel's parameter buffer during compilation.
Then, it performs a byte-pattern match with the known placeholder pointers within the buffer to identify the correct offsets. 
As with our earlier optimization, the transformation remains fully transparent to PyTorch programs and to the rest of the \pt{} compilation framework.

\begin{figure}[t!]
\centering
\begin{minipage}{0.48\linewidth}
\centering
\includegraphics[width=\linewidth]{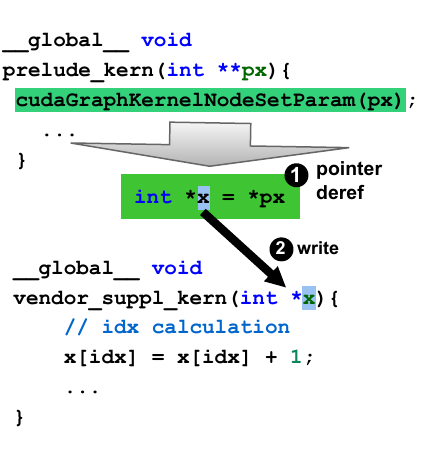}
\caption{Indirection in vendor-supplied kernels.}
\label{fig:non-triton-indirection}
\end{minipage}
\hfill
\begin{minipage}{0.48\linewidth}
\centering
\includegraphics[width=\linewidth]{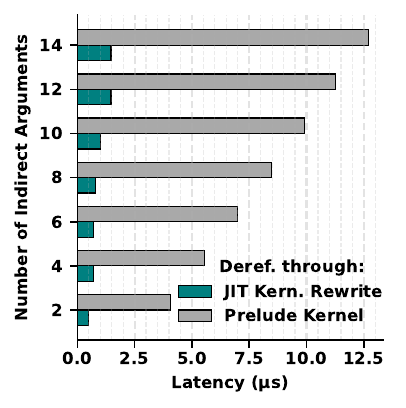}
\caption{Latency comparison of PI approaches.}
\label{fig:triton-non-triton-indirection-overheads}
\end{minipage}
\end{figure}

\autoref{fig:non-triton-indirection} illustrates this technique. 
Recall \tool{}'s aim to convert a kernel parameter (pointer \texttt{x}) into a pointer-to-pointer (\texttt{px}). 
During graph capture (\autoref{imp:graph-capture-layer}), these pointer-to-pointers are recorded as parameters of the prelude kernel. 
Before each replay of the captured \cgacr{}, the invocation-specific pointers to parameters are written into the recorded pointer-to-pointer locations (\texttt{px}). 
During replay, the prelude kernel de-references these pointer-to-pointers (\circled{1} in \autoref{fig:non-triton-indirection}) and writes the resulting values into the parameter fields of the vendor-supplied kernel node (\circled{2}) using the aforementioned parameter update APIs.


A keen reader would note that this technique can also be used to achieve parameter indirection for JIT compiled kernels. 
While true, we empirically observed that it is noticeably slower than rewriting the kernel directly during compilation.
\autoref{fig:triton-non-triton-indirection-overheads} compares the latency of de-reference through JIT kernel rewrite and through prelude kernel techniques as the number of indirected parameters increases. 
\optBacr{} through the prelude kernel is substantially slower and scales poorly, exceeding 10 µs at higher parameter counts. Consequently, we employ JIT-based kernel update where possible, falling back to using prelude kernels only for vendor-supplied kernels. 

\subsubsection{Graph-Capture Layer}
\label{imp:graph-capture-layer}
The \cg{} generator (\mycircled{4} in \autoref{fig:impl:main}) captures the compiled module as a \cgacr{} whenever prior analysis on InductorIR determines that the entire \fg{} can be mapped to a \cgacr{}. 
\pt{} captures \cgacr{}s only at the granularity of a whole-\fg{}. 
Because a \cgacr{} records kernel parameters (\eg{} pointer addresses) by value, the generator creates static placeholders for these parameters and substitutes them for the original parameters before capture.



The graph generator must consider two types of kernel parameters: 
\mycircled{1} those supplied externally (\eg{} application inputs), and
\mycircled{2} those produced by preceding kernels within the \cgacr{}. 
In the former, the placeholders do not initially contain the correct parameter value for each replay.
It must be copied to the placeholders pointed to by the kernel parameters. 
In the second case, no copy is required, since the producer kernel writes directly into the placeholder.
The \cg{} generator thus replaces the original parameters with static placeholders.
It injects code that would perform the necessary parameter copies onto the static placeholder during a kernel's invocation during replay. 
Finally, it uses the stream capture method (\autoref{sec:bg}) to create a \cgacr{} for the module. 

\runheading{\tool{}'s extensions} 
To take advantage of \optBacr{}, \tool{} enhances the \cg{} generator (yellow element in \mycircled{4} in \autoref{fig:impl:main}). 
The enhanced generator receives the list of parameters that had undergone indirection by the code generator. 
For such parameters, \tool{}:
\mycircled{1} allocates static placeholders for \textit{pointers} instead of data regions, and
\mycircled{2} injects code to perform pointer copies for these parameters instead of parameter (data) copies. 
Note that these pointer copies are host-to-device copies over the PCIe, as the list of pointers to parameters originally resides on the CPU.
In contrast, original parameter copies are device-to-device within the GPU. 
This layer also places the prelude kernel into the captured graph.

\subsection{\optC{}}
\label{sec:selector}

Recall that the overheads of a \cg{} can outweigh its benefits (\autoref{sec:motivation}). 
These overheads are due to parameter copies, as well as tasks such as garbage collection for graph replays (\autoref{sec:bg}). 
While \optBacr{} can reduce overheads due to parameter copies for kernels in a \cgacr{}, the rest of the overheads remain. 

\tool{} introduces a profiler in the compilation phase (slow path) to judiciously decide whether to deploy a \cgacr{} based on a cost-benefit analysis (colored green, \mycircled{5} in \autoref{fig:impl:main}). 
The final outcome of the compilation is an optimized executable (module) corresponding to the given block of PyTorch code.
It is then cached for use in the fast path.

\pt{} greedily deploys \cgacr{}s wherever one can be created. 
In contrast, \tool{} considers up to three possible choices: 
\mycircled{1} A module \textit{without} \cgacr{}, 
\mycircled{2} a module with \cgacr{} but \textit{without} \optBacr{}, and 
\mycircled{3} a module with \cgacr{} and \optBacr{}. 
The profiler runs each of the available modules as part of the compilation phase, measures their execution time, and caches the module with the best performance. 

One may wonder why \tool{} distinguishes between \mycircled{2} and \mycircled{3}.
The subtlety is that \optBacr{} replaces device-to-device (D2D) parameter (data) copies with host-to-device (H2D) pointer copies. 
When working with many small objects, it is possible that the overhead of copying pointers over the PCIe outweighs that of copying parameters (data) within the GPU's memory. 
The profiler helps recognize such cases and \tool{} choose to cache the \cgacr{} without \optBacr{}. 
Overall, during compilation, the profiler help \tool{} judiciously choose which \cgacr{}s to deploy during replay for the best performance.

\noindent\underline{\textbf{Summary:}} \tool{} extends \td{} and \ti{} to transparently update IRs for enabling more kernels to execute as part of \cgacr{} (\optAacr{}).
It extends Triton's kernel generator and \pt{}'s \cgacr{} generator to add indirections to convert data copies in \cgacr{} to pointer copies (\optBacr{}). 
Finally, it adds a profiler in the slow path that uses a cost-benefit analysis to decide when it is beneficial to deploy \cgacr{} (\optCacr{}).

\section{Evaluation}
\label{sec:eval}

\begin{figure*}[t!]
    \centering
    \includegraphics[trim={0 10 0 0}, clip, width=\linewidth]{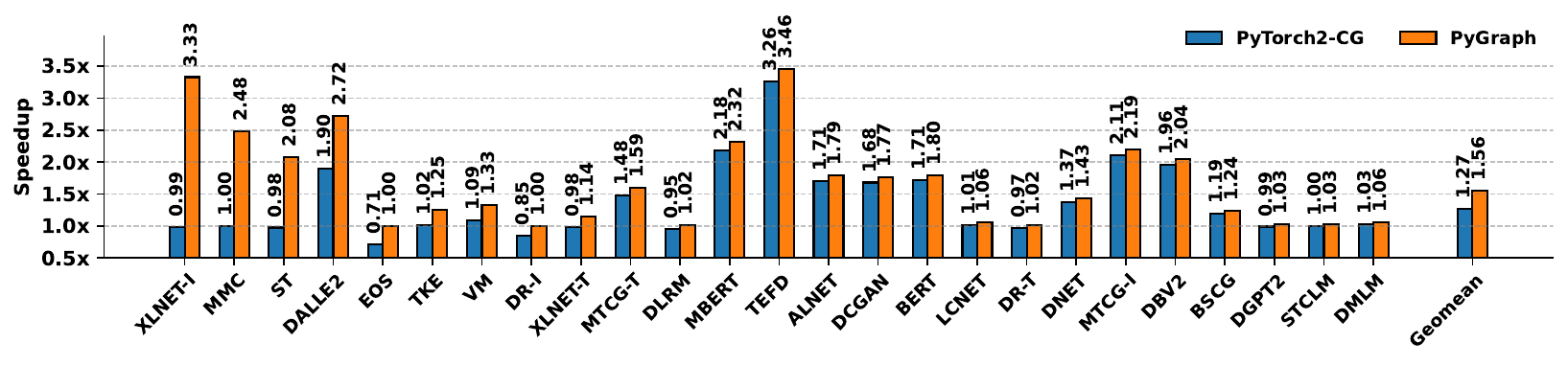}
    \caption{Performance improvement of \ptcg{} and \tool{} against \ptncg{}.}
    \label{fig:eval:main}
\end{figure*}

\begin{table}[]
\scalebox{0.8}{
\begin{tabular}{lp{7.7cm}}
\textbf{Suite} & \textbf{Application (Short Name, Type, Batch Size)} \\ \toprule
TorchBench & Equation of State (EOS, I 1048576), Speech Transformer (ST, I, 1), DALLE2\_pytorch (DALLE2, I, 1), Multi-modal Clip (MMC, I, 1), Turbulent Kinetic Energy (TKE, I, 1048576), Deep Recommender (DR-T, T, 256), Deep Recommender (DR-I, I, 8192), Vision MaskRCNN (VM, I, 1), AlexNet (ALNET, I, 2), dlrm (DLRM, I, 2048), TIMM Efficientdet (TEFD, T, 1), dcgan (DCGAN, T, 32), BERT (BERT, T, 16), densenet121 (DNET, I, 64),   \\ \hline
HuggingFace & XLNet LM-head Model (XLNet-I, I, 1), XLNet LM-head Model (XLNet-T, T, 8), MT5ForConditionalGeneration (MTCG-T, T, 16), MT5ForConditionalGeneration (MTCG-I, I, 16), MobileBertForQuestionAnswering (MBERT, I, 128), DebertaV2ForQuestionAnswering (DBV2, I, 1), BlenderbotSmallForConditionalGeneration (BSCG, T, 64), DistillGPT2 (DGPT2, T, 16), Speech2Text2ForCausalLM (STCLM, T, 256), DebertaForMaskedLM (DMLM, T, 8)   \\ \hline
TIMM & lcnet\_050 (LCNET, I, 256) \\ \bottomrule
\end{tabular}}
\caption{Details of the workloads (T: Training, I: Inference).}
\vspace{-1em}
\label{table:eval:workloads}
\end{table}

Our evaluation seeks to answer the following questions:
\begin{mybullet}
\item How well does \name{} exploit \cg{}s relative to the widely used \pt{} compiler?
\item How much does each optimization in \name{} contribute to end-to-end performance?
\item How robustly does \name{} generalize to distributed models and diverse GPUs?
\end{mybullet}

\noindent
\textbf{Workloads and metric:} 
We evaluate a broad mix of training and inference workloads from TorchBench~\cite{torchbench}, HuggingFace~\cite{hfbench}, and TIMM~\cite{timm}, spanning vision, NLP, and HPC. We focus on applications that are sensitive to \cg{} usage; \autoref{table:eval:workloads} summarizes these workloads. Because ML workloads are inherently repetitive, we use the runtime of a single iteration as our primary metric—one forward pass for inference and one forward–backward pass for training. All reported numbers are averaged over 100 runs.

\noindent
\textbf{Environment:} 
We conduct experiments on an NVIDIA H100 NVL GPU with 94GB HBM3, paired with a 64-core Intel Xeon 8462Y+ CPU and 512 GB of DDR5 memory. The software stack includes PyTorch 2.4~\cite{pytorch2.4}, CUDA 12.8~\cite{cuda-toolkit}, cuDNN 8.9.2~\cite{cudnn8.9.2}, and NVIDIA driver 575.57.08. For multi-GPU benchmarks, we use a system with four 80GB H100s connected via high-bandwidth NVLink, running CUDA 12.4.

\noindent
\textbf{Baselines:} 
We compare \tool{} against two baselines: \pt{} \textit{without} \cg{}s, denoted \ptncg{}, and \pt{} with \cg{}s enabled, denoted \ptcg{}. All compilers are based on PyTorch v2.4. 

\subsection{End-to-end performance}

\begin{figure*}[t!]
    \centering
    \includegraphics[trim={0 10 0 0}, clip, width=\linewidth]{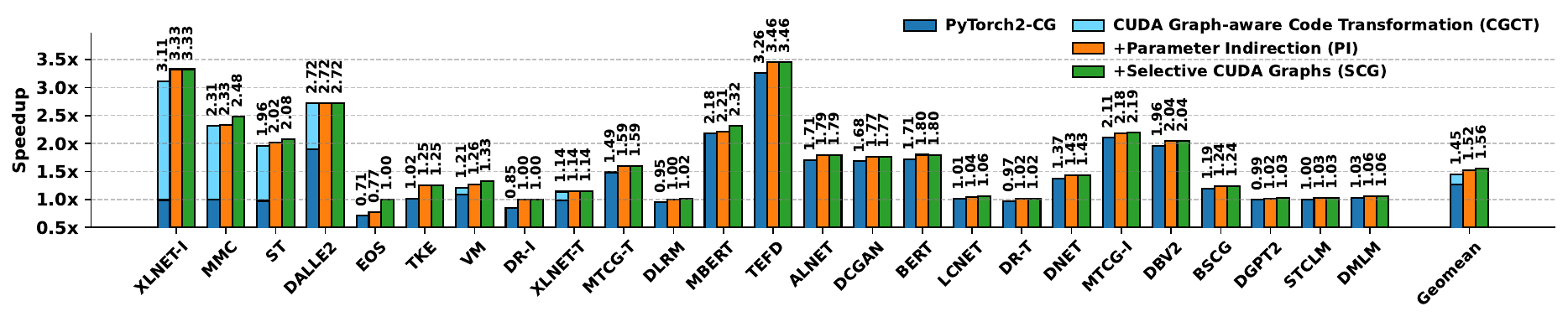}
    \caption{Ablation study on different optimizations of \tool{} (performance normalized against \ptncg{}).}
    \label{fig:eval:ablation:main}
\end{figure*}

\autoref{fig:eval:main} compares \tool{}'s performance against that of \ptncg{} and \ptcg{}.  The figure displays two bars for each application, where the first bar indicates the speedup of \ptcg{} over \ptncg{}, and the second reports the speedup of \tool{} over \ptncg{}. 

First, we find that although \cg{} was introduced purely as a performance optimization, its current deployment in \pt{} can significantly hurt performance. 
\ptcg{} slows down 4 of the 25 applications by at least 3\% compared to \ptncg{} (bars with height $<1$), with worst-case regressions of 29\% for EOS (HPC) and 15\% for Deep Recommender (inference). These workloads consist of short kernels of tens of microseconds each for which the replay overhead (\eg{} parameter copying and RNG management) dominates execution time (often around $\sim50\%$). As a result, blindly deploying all \cgacr{}s causes net slowdowns, and yet \ptcg{} deploys them anyway.
In contrast, \tool{} is at least as good as the best of \ptncg{} and \ptcg{} for every application.  
This is important as it absolves a user from having to decide whether to enable \cgacr{} for each of their programs. 

Second, even in cases where \ptcg{} improves performance over \ptncg{}, it still fails to effectively harness \cg{} as shown by the difference between bars for \ptcg{} and \tool{}. 
\tool{} improves performance over \ptcg{} by 29\% on average and by up to $3.36\times$ (XLNET-inference). 
While \ptcg{} fails to deploy any \cg{}s in XLNET-I, \tool{} deployed $413$ kernels through \cgacr{}s, thanks to its \cgacr{}-aware code transformation.
In another example,
\ptcg{} improves DALLE2 (inference) by $1.9\times$ whereas \name{} improves it by $2.72\times$. 
In summary, \tool{} effectively leverages \cgacr{}s when there is an opportunity without hurting performance elsewhere. 

\subsection{Ablation of individual optimizations}
\label{sec:eval:analysis}
\autoref{fig:eval:ablation:main} quantifies individual contribution of each optimization in \tool{}. 
Each application has three bars corresponding to three optimizations. 
The height of a bar is normalized to the performance of \ptncg{}.
The total height of the first bar shows the speedup obtained by applying \optA{} (\optAacr{}) alone. 
The bottom stack (dark blue) of the first bar shows the speedup of \ptcg{} over \ptncg{}. 
The top stack captures the improvement due to \optAacr{} over \ptcg{}. 
The second bar shows performance of \optBacr{} in tandem with \optAacr{} (+\optB{}). 
The third bar shows the cumulative effect of three optimizations (+\optC{}).  

\noindent\textbf{\optA{}.}  
\optAacr{} enables more kernels to be captured within \cgacr{}s, amortizing the overhead associated with individual kernel launches. 
As a result, applications such as XLNET-I, MMC, and ST achieve speedups of $3.14\times$, $2.31\times$, $2\times$ over \ptcg{}, while DALLE2, XLNET-T, and VM improve performance by $43\%$, $16\%$, and $11\%$ over \ptcg{}, respectively. 
 
\autoref{table:eval:ablation:opt1} lists six applications that witness a significant increase in the percentage of their kernels launched as part of \cgacr{}(s). 
For example, \ptcg{} fails to deploy any \cg{} for MMC and XLNET-I. In case of XLNET-I, its \fg{} with $413$ kernels was not deployed via \cgacr{} due to the presence of a \textit{single} CPU tensor in the \fg{}. 
Similarly, one of the \fg{}s of DALLE2 with $314$ kernels was not deployed as a \cgacr{} due to a \textit{single} memcopy operation from a CPU tensor to a GPU tensor. 
\optAacr{} removes these constraints, deploying more than $99\%$ of the kernels  via \cgacr{}(s).
Even for ST and VM, \optAacr{} increased \cgacr{} coverage from 5.14\% and 58.84\% to 74.22\% and 71.01\%.

\noindent\textbf{\optB{}.} 
\optBacr{} reduces overhead of copying kernel parameters while replaying \cgacr{}s.  
This speeds up TKE by 23\% and DR-I by 18\%.
Twelve others speed up by more than 4\% (EOS, XLNET-I, MTCG-T, TEFD, DCGAN, DLRM, BERT, DR-T, ALNET, DNET, BSCG, VM).

\autoref{tab:eval:ablation:opt2} reports the reduction in bytes copied by \optBacr{}. Most workloads see a drastic drop, with the maximum remaining copy size under \tool{} at just 336 B. For example, it reduces the bytes copied in MTCGT-T, DGPT2, STCLM, and DMLM from 1 GB, 953 MB, 850 MB, and 548 MB, respectively, to just 312 B, 136 B, 136 B, and 136 B, a reduction of over 99\%. 
\autoref{tab:eval:ablation:opt2} also shows the number of \cgacr{}s that \tool{} optimizes via \optBacr{} in parenthesis. 

Note that even after eliminating all data copy, \optBacr{} may not always benefit the application. 
For example, for a small tensor, the PCIe transfer cost due to pointer copy may outweigh the benefit of avoiding a larger copy within GPU HBM. These are the cases where a \cgacr{} does not benefit from \optBacr{}. 
For instance, DALLE2 sees no benefit from \optBacr{}.
In contrast, the lone \cgacr{} in TKE and DR-I gains significantly from reduced copy sizes, yielding speedups of $23\%$ and $18\%$.

\noindent\textbf{\optC{}.}
\optCacr{} judiciously deploys \cgacr{}s driven by a cost-benefit analysis. 
It disables \cgacr{}s where they could hurt performance, avoiding regression.
This is particularly important for applications such as EOS, where blind deployment of \cgacr{}s  degrades performance by $29\%$.

\begin{table}[t!]
\centering
\renewcommand{\arraystretch}{1.1}
\scriptsize
\setlength{\tabcolsep}{4pt}
\begin{tabular}{|l|c|c|}
\hline
\multirow{2}{*}{\textbf{App.}} & \multicolumn{2}{c|}{\textbf{\% of Kernels in \cg{}s}} \\ \cline{2-3} 
                                & \textbf{\ptcg{}}                & \textbf{w/ \optA{}}            \\ \hline
ST                            & 5.14                             & 74.22                       \\ \hline
DALLE2                        & 79.56                            & 99.32                       \\ \hline
MMC                           & 0                            & 99.32                         \\ \hline
XLNET-T                       & 1.53                             & 98.06                          \\ \hline
XLNET-I                       & 0                             & 99.28                        \\ \hline
VM                            & 58.84                            & 71.01                        \\ \hline
\end{tabular}
\caption{Enhanced coverage by \cg{}s due to \optA{}.} 
\label{table:eval:ablation:opt1}
\end{table}

\begin{table}[t!]
\centering
\renewcommand{\arraystretch}{1.1}
\scriptsize
\setlength{\tabcolsep}{4pt}
\begin{tabular}{|l|c||c|c|}
\hline
\textbf{App.} & \begin{tabular}[c]{@{}c@{}} \textbf{Bytes Copy Reduced}\\ (\textbf{\# \cg{}s})\end{tabular} & \textbf{App.} & \begin{tabular}[c]{@{}c@{}}\textbf{Bytes Copy Reduced}\\ (\textbf{\# \cg{}s})\end{tabular} \\ \hline
XLNET-I       & 8.0  KB $\rightarrow$ 16.0 B   (1)           & MMC       & 314.0 KB $\rightarrow$ 16.0 B   (1) \\ \hline
ST            & 1.3 MB $\rightarrow$ 336.0 B   (3)           & DALLE2    & 4.0 KB $\rightarrow$ 16.0 B   (5) \\ \hline
EOS           & 4.1 MB $\rightarrow$ 24.0 B   (0)           & TKE       & 174.0 MB $\rightarrow$ 168.0 B   (1) \\ \hline
VM            & 1.6 MB $\rightarrow$ 184.0 B   (3)           & DR-I      & 3.0 GB $\rightarrow$ 8.0 B   (1) \\ \hline
XLNET-T       & 362.0 MB $\rightarrow$ 232.0 B   (3)           & MTCG-T    & 1.0 GB $\rightarrow$ 312.0 B   (3) \\ \hline
DLRM          & 14.6 MB $\rightarrow$ 136.0 B   (1)           & MBERT     & 130.0 KB $\rightarrow$ 24.0 B   (1) \\ \hline
TEFD          & 6.2 MB $\rightarrow$ 120.0 B   (2)           & ALNET     & 588.0 KB $\rightarrow$ 8.0 B   (1) \\ \hline
DCGAN         & 1.5 MB $\rightarrow$ 16.0 B   (1)           & BERT      & 156.3 MB $\rightarrow$ 32.0 B  (2) \\ \hline
LCNET         & 73.5 MB $\rightarrow$ 8.0 B   (1)           & DR-T      & 290.0 MB $\rightarrow$ 16.0 B   (2) \\ \hline
DNET          & 18.4 MB $\rightarrow$ 8.0 B   (1)           & MTCG-I    & 32.0 KB $\rightarrow$ 16.0 B   (1) \\ \hline
DBV2          & 4.0 KB $\rightarrow$ 24.0 B   (1)           & BSCG      & 1.6 GB $\rightarrow$ 56.0 B   (30) \\ \hline
DGPT2         & 953.4 MB $\rightarrow$ 136.0 B   (3)           & STCLM     & 849.5 MB $\rightarrow$ 136.0 B   (14) \\ \hline
DMLM          & 548.8 MB $\rightarrow$ 136.0 B   (3)           &           &                                \\ \hline
\end{tabular}
\caption{Reduction in data copy and number of \cg{}s optimized by \tool{}'s parameter indirection.}
\vspace{-1em}
\label{tab:eval:ablation:opt2}
\end{table}

In five applications, \tool{} selectively disables \cgacr{}s to achieve better overall performance. 
For instance, VM exposes 21 candidate \cgacr{}s, but \name{} enables only four and disables the other 17 to improve end-to-end performance by 6\%.  
Similarly, \optCacr{} disables the single \cgacr{} in EOS to avoid performance loss. 
Overall, \tool{} exercises selective deployment by enabling 97 of 123 possible \cgacr{}s across the applications.


Overall, these results demonstrate that each of \tool{}’s optimizations makes a meaningful contribution across multiple applications, collectively enhancing overall performance.

\subsection{Evaluating distributed ML models}
\begin{figure}[tb!]
\centering
\includegraphics[width=\linewidth]{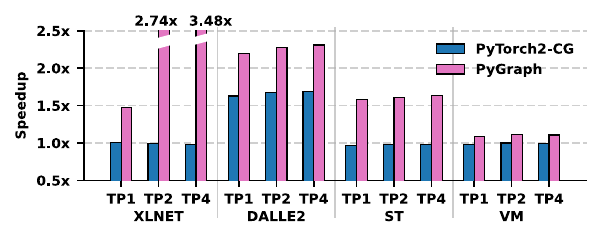}
\caption{Speedup of \ptcg{} and \tool{} over \ptncg{} across tensor parallel settings (TP=1,2,4).}
\label{fig:TP-speedup}
\vspace{-1.5em}
\end{figure}

We evaluated the efficacy of \tool{} under tensor parallelism (TP)~\cite{megatron-lm}, a widely deployed strategy for distributing ML processing in a multi-GPU setting.  
We evaluated four workloads under TP on server with 4$\times$H100 GPUs connected over NVLink. 
\autoref{fig:TP-speedup} summarizes the results: for each application, we report three groups of measurements corresponding to TP-1, TP-2, and TP-4. 
In each group, we compare the performance of \ptcg{} and \name{} against the \ptncg{} under the given TP setting. This allows us to observe the effectiveness of \ptcg{} and \name{} under TP.

\autoref{fig:TP-speedup} demonstrates that \name{} significantly amplifies the benefits of \cgacr{}s over \ptcg{} under TP. 
All results use inference batch size 1, except for XLNET, which uses batch size 128.
As the degree of TP increases, GPU kernels finish faster, and CPU launch latency becomes a larger fraction of overall runtime. 
This, in turn, increases the importance of fully harnessing \cg{}s, which \ptcg{} fails to do. 
In contrast,  \name{} yields substantially higher speedups for XLNET, DALLE2, and ST with progressively larger gains as parallelism increases, as it is able to deploy many kernels in \cg{}s.
Its speedup reaches up to $3.48\times$ over \ptncg{} at TP-4 for XLNET with $2.41\times$ on average across different TP settings. 
This demonstrates that \tool{} assumes further importance with increasingly wider deployment of distributed ML processing across multiple GPUs.

\noindent
\textbf{Evaluation on A6000 GPU.} We also evaluated \name{}  on a workstation class A6000 GPU having less compute power than H100s. 
As the GPU becomes slower, the headroom for improvement with \cgacr{}s reduces. 
Even then, \tool{} more prudently utilizes \cg{} compared to \ptcg{}.
For example, \ptcg{} degrades performance compared to \ptncg{} by up to $32\%$ for EOS. In contrast, \name{} never degrades performance and achieves a maximum improvement of $3.25\times$ over \ptncg{} with an average speedup of $1.18\times$ compared to \ptcg{}'s average speedup of $1.06$ over \ptncg{}. This shows that \name{}'s efficacy is not tied to a specific GPU architecture.

\noindent
\textbf{Compilation Time.} Across all the workloads we evaluated, \name{} increases compilation time relative to PyTorch2 by $1.6\times$ on average (\eg{} compilation time for ST and DALLE2 increased from 15 seconds and 96 seconds to 37 seconds and 185 seconds, respectively). This extra latency reflects the additional work \name{} performs, such as code transformations, kernel rewriting, and profiling. However, note that compilation is a one-time cost incurred in the slow path that is easily covered by the steady-state performance improvement in the fast path.

\section{Related Work}
\label{sec:rel}

\noindent\textbf{ML compilers:}
Early ML compilers such as TensorFlow~\cite{tensorflow}, Caffe~\cite{caffe}, Theano~\cite{theano}, and CNTK~\cite{cntk} improve performance via graph-level optimizations, while later systems develop richer tensor-program IRs~\cite{tiramisu,tvm,ding:hidel:23,tensorir,Graphene,program-in-halide,shao2022tensor,triton,unit,bolt,akg,ansor,amos,flextensor,roller} and target efficient code generation across diverse hardware and irregular workloads~\cite{Shen2020NimbleEC,coratensorcompilercompilation,freetensor,sparsetir}. 
Recent advances introduce whole-graph and memory-centric optimizations~\cite{welder,souffle}, accelerate dynamic control flow~\cite{cocktailer}, and perform multi-level super-optimization using unified IRs~\cite{mirage}. 
These efforts reduce kernel count and/or improve efficiency of kernels.
\tool{} complements them by mitigating CPU launch overhead and enhancing \cg{}'s deployability. 

\pt{}~\cite{pytorch2} showed that dynamic Python execution can coexist with graph compilation by using \td{}~\cite{td} to rewrite Python bytecode into FxGraphs~\cite{torchfx} and \ti{}~\cite{ti} to lower them into optimized GPU kernels. 
Earlier capture mechanisms--JIT tracing, TorchScript~\cite{torchscript}, lazy tensors~\cite{pytorch-xla,lazytensor}, and symbolic tracing~\cite{torchfx}--were limited by unsoundness, incomplete Python coverage, or high recompilation overheads. 
\tool{} is not tied to \pt{}’s pipeline; it only assumes a compiler that produces a stable kernel-launch DAG and augments the backend that prepares and executes kernels.
\tool{}'s approach can integrated with JAX~\cite{jax}, XLA~\cite{xla}, too. 
Source-level graph-repair techniques~\cite{graphmend} are complementary to \tool{}. 
They lengthen computation graphs, while \tool{} ensures the resulting kernel DAGs are effectively captured as \cg{}s.

\noindent\textbf{\cg{}s:}
Grape~\cite{zheng:grape:23} is a close prior work, optimizing \cg{}s for DNNs through predication rewriting, metadata compression, and alias prediction, but these techniques require extensive manual kernel and CUDA-driver changes -- \eg{} over 400 lines of kernel edits for GPT beam search -- and its alias predictor is incompatible with \pt{}. 
In contrast, \tool{} requires \textit{no} manual modifications and works transparently for all PyTorch applications. 
Extensive manual changes required by Grape and its tight integration with PyTorch 1.12 makes a quantitative comparison infeasible.
Other systems use \cg{}s in specialized deployments, such as serverless LLM inference, by materializing and restoring pre-captured graphs~\cite{medusa}. 
These efforts focus on a graph's persistence, whereas \tool{} enables broader and more efficient deployment of \cg{} and thus, is complementary.

\section{Conclusion}
\label{sec:conclusion}



We present \tool{} — a compiler framework designed to enhance the deployment and performance of \cg{}s. 
\tool{} addresses several key limitations of the \pt{} compiler, leading to substantial performance improvement across various ML benchmarks without requiring manual modifications to applications.


\bibliographystyle{plain}
\bibliography{references}

\end{document}